%% file: acl_latex.tex
\pdfoutput=1

\documentclass[11pt]{article}

\usepackage{acl}

\usepackage{times}
\usepackage{latexsym}

\usepackage[T1]{fontenc}

\usepackage[utf8]{inputenc}

\usepackage{microtype}

\title{InstructEval: Systematic Evaluation of Instruction Selection Methods}

\author{Anirudh Ajith$^{*}$ \quad Chris Pan$^{*}$\quad Mengzhou Xia \quad Ameet Deshpande \quad Karthik Narasimhan \\
\textnormal{Department of Computer Science, Princeton University} \\
\texttt{\{anirudh.ajith, chrispan, mengzhou, asd, karthikn\}@princeton.edu}
}

\input{packages_and_commands}

\begin{document}
\maketitle
\renewcommand{\thefootnote}{\fnsymbol{footnote}}
\footnotetext[1]{Equal contribution}
\renewcommand{\thefootnote}{\arabic{footnote}}

\input{sections/abstract}
\input{sections/introduction}
\input{sections/related_work}
\input{sections/evaluation_suite}
\input{sections/experimental_setup}

\input{sections/results}
\input{sections/conclusion}

\bibliography{anthology,custom}
\bibliographystyle{acl_natbib}

\input{sections/appendix}
\end{document}

%% file: packages_and_commands.tex
\usepackage{amsmath,amsfonts}
\usepackage{booktabs,arydshln}
\usepackage{multirow}
\usepackage{ulem}
\usepackage{pifont}
\usepackage{relsize}
\usepackage{enumitem}
\usepackage{graphicx}
\usepackage{amssymb}
\usepackage{pifont}
\usepackage{dcolumn}
\usepackage{xcolor,colortbl}
\usepackage{supertabular}
\usepackage{tikz}
\usepackage{subcaption}
\usepackage{svg}
\usepackage{float}
\restylefloat{table}
\usepackage{enumitem}

\newcolumntype{d}[1]{D{.}{.}{#1}}  

\newcommand{\cmark}{\ding{51}}%
\newcommand{\xmark}{\ding{55}}%
\newcommand{\greyrule}{\arrayrulecolor{black!30}\midrule\arrayrulecolor{black}}

\makeatletter
\def\adl@drawiv#1#2#3{%
        \hskip.5\tabcolsep
        \xleaders#3{#2.5\@tempdimb #1{1}#2.5\@tempdimb}%
                #2\z@ plus1fil minus1fil\relax
        \hskip.5\tabcolsep}
\newcommand{\cdashlinelr}[1]{%
  \noalign{\vskip\aboverulesep
           \global\let\@dashdrawstore\adl@draw
           \global\let\adl@draw\adl@drawiv}
  \cdashline{#1}
  \noalign{\global\let\adl@draw\@dashdrawstore
           \vskip\belowrulesep}}
\makeatother

\newcommand{\secref}[1]{Section~\ref{#1}}

\newcommand{\tableref}[1]{Table~\ref{#1}}

\usetikzlibrary{backgrounds, shapes,arrows,positioning,matrix,fit}

\definecolor{lightblue}{RGB}{173,216,230}
\definecolor{lightpink}{RGB}{255,182,193}
\definecolor{lightyellow}{rgb}{0.9,0.9,0.6}
\definecolor{darkyellow}{rgb}{0.7, 0.7, 0.3}

\pgfdeclarelayer{main}
\pgfdeclarelayer{layer1}
\pgfdeclarelayer{layer2}
\pgfdeclarelayer{layer3}
\pgfdeclarelayer{layer4}
\pgfdeclarelayer{layer5}
\pgfdeclarelayer{layer6}
\pgfsetlayers{layer6, layer5,layer4,layer3, layer2, layer1, main}

%% file: sections/abstract.tex
\begin{abstract}

In-context learning (ICL) performs tasks by prompting a large language model (LLM) using an instruction and a small set of annotated examples called demonstrations.
Recent work has shown that precise details of the inputs used in the ICL prompt significantly impact performance, which has incentivized instruction selection algorithms.
The effect of instruction-choice however is severely underexplored, with existing analyses restricted to shallow subsets of models and tasks, limiting the generalizability of their insights.
We develop InstructEval, an ICL evaluation suite to conduct a thorough assessment of these techniques. The suite includes 13 open-sourced LLMs of varying scales from four model families, and covers nine tasks across three categories.
Using the suite, we evaluate the relative performance of seven popular instruction selection methods over five metrics relevant to ICL.
Our experiments reveal that using curated manually-written instructions or simple instructions without any task-specific descriptions often elicits superior ICL performance overall than that of automatic instruction-induction methods, pointing to a lack of generalizability among the latter.
We release our evaluation suite for benchmarking instruction selection approaches and enabling more generalizable methods in this space.\footnote{Code: \url{https://github.com/princeton-nlp/InstructEval}}

\end{abstract}

%% file: sections/introduction.tex
\section{Introduction}

\begin{figure}[t]
    \centering
    \includegraphics[width=\linewidth]{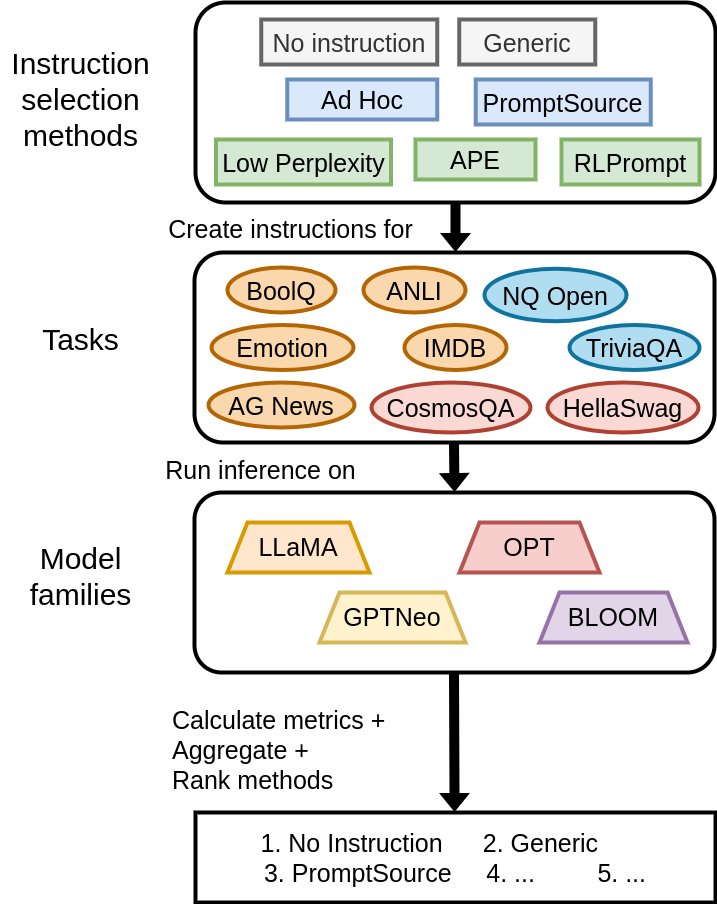}
    \caption{
        InstructEval allows the assessment of instruction selection methods for ICL across a range of models and tasks along five metrics.
     }
    \label{fig:teaser}
\end{figure}

One of the most effective insights in NLP research in recent years has been that large language models trained to perform next-token prediction show emergent in-context learning (ICL) abilities~\cite{brown2020language, Scao2022BLOOMA1, Zhang2022OPTOP}.
While the bulk of research interest has shifted away from task-specific models and towards creating ``foundation models" to perform a variety of tasks using appropriately constructed prompts, the performance of ICL remains sensitive to the precise details of prompt construction. Prompt engineering remains critical for achieving optimal ICL performance~\cite{perez2021true, zhao2021calibrate, webson-pavlick-2022-prompt}.

In practice, ICL typically involves prompting a language model using a concatenation of a task-specific \textit{instruction}, a short sequence of annotated in-context examples known as \textit{demonstrations}, and a \textit{test example} (Figure \ref{fig:prompt_example}).
Much of the research interest surrounding in-context learning has focused on understanding the optimal selection, ordering of demonstrations, and label-space choices~\cite{Liu2021WhatMG, Su2022SelectiveAM, rubin2022learning, Wang2023LargeLM, lu2021fantastically, Wei2023LargerLM, Pan2023WhatIL}. However, \emph{instruction choice} remains a relatively underexplored aspect of prompt engineering despite its established significance~\cite{mishra-etal-2022-cross} on downstream performance.

Even among recent works exploring automatic instruction selection~\cite{honovich2022instruction, lowperplexityprompts, rlprompt, ape}, the use of different evaluation protocols makes the comparison of their relative performances difficult. Existing studies typically limit their analyses to specific models or tasks; for example, \citet{ape} focus on a single model, and while \citet{rlprompt} consider multiple model scales, they all belong to a single model family. Moreover, evaluations often span disparate task selections with minimal overlap and are primarily dominated by classification tasks, neglecting other task types like multiple-choice QA or generation. Lastly, most previous works tend to emphasize zero-shot accuracy, overlooking other pertinent ICL metrics such as few-shot accuracy and robustness measures.

To address these issues, we build InstructEval, an evaluation suite for the comprehensive evaluation of instruction selection methods. The suite covers a diverse collection of 13 open-sourced autoregressive LLMs from four model families and nine tasks spanning three task types. Additionally, it also incorporates three accuracy metrics and two sensitivity metrics that are of interest to ICL.
We perform evaluations of seven popular instruction selection methods including trivial instruction baselines, manually curated instructions, and sophisticated automatic methods using our suite. 

Overall, we find that the relative effectiveness of these approaches varies significantly across different models and task types. We discover that curated manually-written instructions and task-agnostic instructions can elicit better aggregated performance (over models) than automatically induced ones, highlighting the lack of generalizability of the latter. We also find that including instructions in few-shot prompts usually tends to hurt ICL performance at the model scales we consider. Our findings suggest that it may be optimal for ICL practitioners to omit instructions in few-shot settings and use curated manually-written instructions in zero-shot settings, rather than contemporary automatic induction techniques that require substantial computation and hyperparameter tuning to achieve competitive performance.
We release the evaluation suite we develop to aid the systematic study of even more questions regarding prompt engineering that we do not explicitly address in our work.

%% file: sections/related_work.tex
\section{Related Work}

\paragraph{In-Context Learning and Existing Benchmarks} As language models have scaled, in-context learning has emerged as a popular paradigm and remains ubiquitous among several autoregressive LLM families \cite{brown2020language, llama, bloom, gptneo, opt}. Benchmarks like BigBench \cite{bigbench} and HELM \cite{helm} have been created for the holistic evaluation of these models. BigBench focuses on few-shot abilities of state-of-the-art large language models, while HELM extends to consider metrics like robustness and bias. However, these benchmarks focus on evaluating and ranking \emph{language models}, and do not address the systematic evaluation of \emph{prompting methods}. Although contemporary work by \citet{yang2023improving} also aims to perform a similar systematic analysis of prompting methods, they focus on simple probability-based prompt selection while we evaluate a broader range of methods including trivial instruction baselines, curated manually selected instructions, and sophisticated automated instruction selection.

\paragraph{Automated Prompt Engineering Methods} There has been interest in performing automated prompt-engineering for target downstream tasks within ICL. This has led to the exploration of various prompting methods, ranging from simple heuristics such as selecting instructions with the lowest perplexity \cite{lowperplexityprompts}, inducing instructions from large language models using a few annotated input-output pairs \cite{ape}, to utilizing RL objectives to create discrete token sequences as prompts \cite{rlprompt}. However, these works restrict their evaluation to small sets of models and tasks with little intersection, hindering their objective comparison. 

\paragraph{Understanding in-context learning} There has been much recent work attempting to understand the mechanisms that drive in-context learning. Studies have found that the selection of demonstrations included in prompts significantly impacts few-shot accuracy across most tasks \cite{whatmakesgoodicexamples, selectionmachinetranslation, knnprompting}. Works like \cite{fantasticallyorderedprompts} also show that altering the ordering of a fixed set of demonstrations can affect downstream accuracy. Prompts sensitive to demonstration permutation often exhibit lower accuracies \cite{relationsensitivityaccuracy}, making them less reliable, particularly in low-resource domains.

Our work aims to bridge these gaps by systematically evaluating the efficacy of popular instruction selection approaches over a diverse set of tasks and models, facilitating objective comparison. We evaluate these methods not only on accuracy metrics, but also on sensitivity metrics to glean additional insights. We recognize that other facets of prompting not covered by instruction engineering exist \cite{weichain, react, selfconsistency}, and defer these explorations to future work. 

%% file: sections/evaluation_suite.tex
\section{Evaluation Suite}

\subsection{Prompt format}

\begin{figure}[t]
    \centering
    \includegraphics[width=\linewidth]{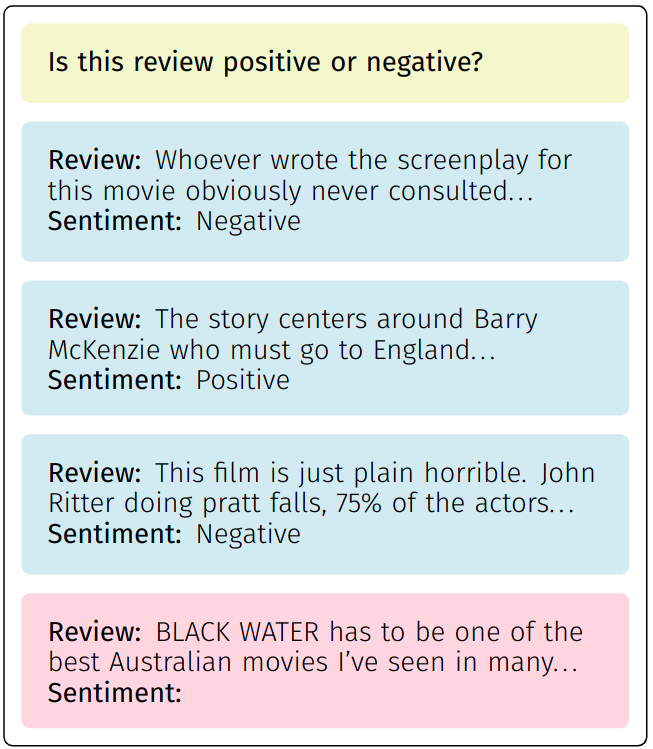}
    \caption{
        An example of a prompt following the template we use for IMDB. By  `prompt' we refer to the concatenation of the \textcolor{darkyellow}{instruction}, solved \textcolor{blue}{demonstrations} and an unsolved \textcolor{red}{test example}.
    }
    \label{fig:prompt_example}
\end{figure}
We define a `prompt' as the full textual input provided to an LLM. Our evaluation suite supports the use of any number of demonstrations, arbitrary demonstration templates and the inclusion of custom strings anywhere within the prompt. Since the instructions used can be set to any arbitrary strings, users are free to use any external means to select instructions and have them evaluated by our suite.

For consistency, we conduct all experiments in this work using prompts that begin with an instruction, continue with a sequence of annotated training demonstrations, and conclude with an unsolved test example\footnote{Instructions are omitted during `Null instruction' evaluations. Demonstrations are omitted in zero-shot evaluations.} (Figure~\ref{fig:prompt_example}), and express each example in a minimal, task-specific key-value format (Table~\ref{table:templates}) that reflects task semantics.

\subsection{Metrics}
\begin{figure*}
  \begin{subfigure}[b]{0.32\textwidth}
    \centering
    \resizebox{\linewidth}{!}{
    \input{figures/perturbation_example_tikz}
    }
    \caption{Perturbation accuracy}
  \end{subfigure}\hfill
  \begin{subfigure}[b]{0.32\textwidth}
    \centering
    \resizebox{\linewidth}{!}{
    \input{figures/selection_example_tikz}
    }
    \caption{Selectional sensitivity}
  \end{subfigure}\hfill
  \begin{subfigure}[b]{0.32\textwidth}
    \centering
    \resizebox{\linewidth}{!}{
    \input{figures/permutation_example_tikz}
    }
    \caption{Permutational sensitivity}
  \end{subfigure}
  \caption{
        We provide schematic diagrams that show prompts are modified to measure \textit{perturbation accuracy}, \textit{selectional sensitivity} and \textit{permutational sensitivity}. We perturb the test input to measure perturbation accuracy, and demonstration selection and permutation respectively while measuring selectional and permutational sensitivity.
    }
    \label{fig:sensitivity-metrics}
\end{figure*}
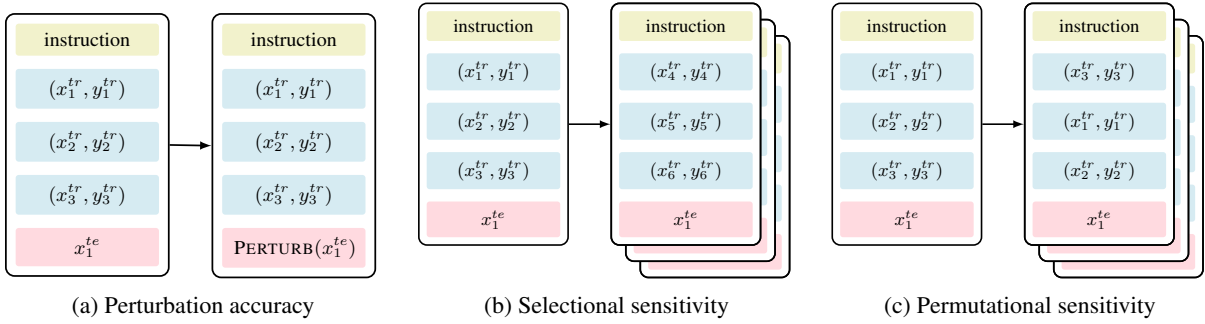

\label{sec:metrics}
\paragraph{Accuracy metrics} Accuracy is typically the primary metric of interest in ICL. While ICL is most commonly performed in few-shot settings where a handful of annotated demonstrations are included in the prompt, models are also prompted zero-shot without the use of such demonstrations. Since real-world scenarios can often contain grammatical errors and misspellings in the test input, it is desirable to find prompts robust to these perturbations. Hence, we measure \textit{zero-shot accuracy}, \textit{few-shot accuracy}, and \textit{perturbation accuracy}\footnote{We choose to treat this as an accuracy metric rather than a sensitivity metric since it is not meaningful to measure sensitivity to such perturbations in situations where a prompt only elicits near random-chance task performance from a model.} in our evaluations. Following \citet{helm}, we measure perturbation accuracy by introducing random capitalization, spacing, contractions and common misspellings in the test input.

\paragraph{Sensitivity metrics} Previous work has shown that the accuracy obtained using a prompt template can fluctuate significantly as a function of the set of demonstrations included in the prompt ~\cite{Liu2021WhatMG, Su2022SelectiveAM, rubin2022learning, Wang2023LargeLM} and the order they are presented in \cite{fantasticallyorderedprompts}. It may be desirable in practice to identify prompt templates and instructions that offer consistent performance regardless of the choice of demonstrations and their arrangement. Hence, we introduce \textit{selectional sensitivity} and \textit{permutational sensitivity} metrics to measure the sensitivity of chosen instructions respectively to selected demonstrations, and the order in which they are arranged. We quantify the sensitivity of an instruction (given a model and task) using the standard deviation of accuracies obtained on varying  the selection or permutation of the demonstrations used, each across 16 random choices.

\subsection{Aggregating metrics across Models}

Each instruction selection method being tested across $N$ models and $M$ datasets yields $NM$ values per metric. Comparing these $NM$-dimensional vectors directly is complex. It can be challenging to reduce them to a single representative scalar. Simple approaches such as computing the mean of these $NM$ values can prove inadequate since the resulting scores would tend to be heavily influenced by metric values that exhibit a high variance across different inspected methods. 

We opt against using aggregation techniques used by previous works \cite{helm, bigbench} due to their drawbacks (Section~\ref{app:scoring}) and instead adopt `mean relative gain' as a means to aggregate accuracy metrics across multiple models. We rely on simple averaging for sensitivity metrics, partly because we observe that these quantities do not show much variation across methods.

\subsubsection{Accuracy metrics}
Considering the range of models and datasets in our evaluation suite, we unsurprisingly observe substantial variation in accuracy magnitudes across model scales and tasks. However, we notice that the degree of variation in accuracy due to instruction choice is usually considerably smaller than the degree of variation due to model and task choice.

To meaningfully compare and aggregate the relative performance of different instruction selection methods across models, we use a measure called \textit{mean relative gain}. First, we define the \textit{relative gain} for a value $x$ from a population $P$ as the percentage by which $x$ exceeds the mean value of $P$:

$$\text{r-gain}_P(x) = 100 \times \dfrac{x-\mu_P}{\mu_P}$$

Consider a collection of models $\mathcal{M}$ and instructions $\mathcal{I}$ for a task $t$. Given a model $m$, we calculate the raw accuracy scores $s_{tmi}$ for each instruction $i \in \mathcal{I}$. Taking this set $S_{tm}$ to be the population, we compare the performances of the instructions against each other by computing their corresponding relative gains $r_{tmi} = \text{r-gain}_{S_{tm}}(s_{tmi})$. Each $r_{tmi}$ represents the degree by which method $i$ outperforms the average performance along the metric on task $t$ for model $m$.

We now define the mean relative gain as 
$$\overline{r}_{ti} = \dfrac{1}{|\mathcal{M}|} \sum_{m \in \mathcal{M}} r_{tmi}$$

These $\overline{r}_{ti}$ values, tabulated and analyzed in \secref{sec:results}, capture not only the ordinal information about each method's performance on a given task but also provide an intuitive sense of the magnitude by which these methods outperform others. Specifically, if an induction method $i$ has a mean relative gain $\overline{r}_{ti}$ on task $t$, this means that method $i$ exceeds average performance (across $\mathcal{I}$) on task $t$ by $\overline{r}_{ti}$ percent when averaged across models $\mathcal{M}$. 

\subsubsection{Sensitivity metrics}
To aggregate the sensitivity of an instruction selection/induction method $i$ over all models for a task $t$, we simply compute the average of the raw sensitivity scores (described in \secref{sec:metrics}). Specifically, if $\sigma_{tmi}$ is the raw sensitivity score obtained for model $m$ and task $t$ when using instruction $i$, then the aggregated sensitivity score $\overline{\sigma}_{ti}$ is given by 

$$\overline{\sigma}_{ti} = \dfrac{1}{|\mathcal{M}|} \sum_{m \in \mathcal{M}} \sigma_{tmi}$$

We choose to avoid more sophisticated aggregation strategies like relative gain for sensitivity metrics since standard deviations are already secondary metrics making it unintuitive to discuss the relative gain of the standard deviation obtained using a method over the average. 

\subsection{Tasks}
\input{tables/tasks} 
While previous instruction induction \cite{ape, rlprompt} work has tended to focus mostly on classification tasks, we include 9 tasks (\tableref{table:tasks}) in our evaluation suite spanning classification (CLS), multiple-choice question-answering (MCQ) and generative question-answering (GQA) to assess the applicability of instruction selection and induction methods to other task-types as well. We concentrate on tasks that are challenging to contemporary language models, and yet are not so demanding that the performance of these models does not exceed random chance. We exclude certain generative tasks, like summarization, which are challenging to assess objectively.~\footnote{Standard summarization metrics correlate poorly with human preferences~\cite{helm, goyal2023news}.}

\input{tables/models}

\subsection{Models}
\input{tables/methods}
We include a diverse range of 13 autoregressive LLMs (\tableref{table:models}) from 4 model families of sizes ranging from 1.1 billion to 20 billion parameters in our evaluation suite. We choose contemporary models that span different architectures and training paradigms which are known to show good ICL performance. This diversity bolsters the generalizability of insights obtained using our evaluation suite while mitigating potential bias towards any specific model family. Moreover, we select open-source models which are large enough to show non-trivial ICL performance while still being small enough to run on reasonable consumer hardware to ensure the practical significance of our findings.

%% file: figures/perturbation_example_tikz.tex
\begin{tikzpicture}[
  node distance = 0.2cm and 1cm, 
  single node/.style = {rectangle, fill=#1, text opacity=1, opacity=0.5, align=center, font=\footnotesize, text width=2cm, inner sep=1.5mm, rounded corners=0.5mm},
  bigbox/.style = {draw, thick, rounded corners, rectangle, inner sep=1.5mm}, 
  arrow/.style={thick, -latex},
]

\node[single node=lightyellow] (instruction1) {instruction};
\node[single node=lightblue, below=of instruction1] (example1) {$(x_1^{tr}, y_1^{tr})$};
\node[single node=lightblue, below=of example1] (example2) {$(x_2^{tr}, y_2^{tr})$};
\node[single node=lightblue, below=of example2] (example3) {$(x_3^{tr}, y_3^{tr})$};
\node[single node=lightpink, below=of example3] (test1) {$x_1^{te}$};

\node[bigbox, fit=(instruction1) (example1) (example2) (example3) (test1)] (box1) {};

\node[single node=lightyellow, right=of instruction1] (instruction2) {instruction};
\node[single node=lightblue, below=of instruction2] (example12) {$(x_1^{tr}, y_1^{tr})$};
\node[single node=lightblue, below=of example12] (example22) {$(x_2^{tr}, y_2^{tr})$};
\node[single node=lightblue, below=of example22] (example32) {$(x_3^{tr}, y_3^{tr})$};
\node[single node=lightpink, below=of example32] (test2) {\textsc{Perturb}$(x_1^{te})$};

\node[bigbox, fit=(instruction2) (example12) (example22) (example32) (test2)] (box2) {};

\draw [arrow] (box1.east) -- (box2.west);
\end{tikzpicture}

%% file: figures/selection_example_tikz.tex
\begin{tikzpicture}[
  node distance = 0.2cm and 0.2cm,
  single node/.style = {rectangle, fill=#1, text opacity=1, opacity=0.5, align=center, font=\footnotesize, text width=2cm, inner sep=1.5mm, rounded corners=0.5mm},
  bigbox/.style = {draw, thick, rounded corners, rectangle},
  arrow/.style={thick, -latex},
  bigbox opaque/.style = {draw, thick, rounded corners, rectangle, opacity=1, fill=white},
  arrow/.style={thick, -latex},
]

\node[single node=lightyellow] (instruction) {instruction};

\node[single node=lightblue, below=of instruction] (example1) {$(x_1^{tr}, y_1^{tr})$};
\node[single node=lightblue, below=of example1] (example2) {$(x_2^{tr}, y_2^{tr})$};
\node[single node=lightblue, below=of example2] (example3) {$(x_3^{tr}, y_3^{tr})$};

\node[single node=lightpink, below=of example3] (test) {$x_1^{te}$};

\node[bigbox, fit=(instruction) (example1) (example2) (example3) (test)] (box1) {};

\begin{pgfonlayer}{layer1}
\node[single node=lightyellow, right=1cm of instruction] (instruction2) {instruction};

\node[single node=lightblue, below=of instruction2] (example12) {$(x_4^{tr}, y_4^{tr})$};
\node[single node=lightblue, below=of example12] (example22) {$(x_5^{tr}, y_5^{tr})$};
\node[single node=lightblue, below=of example22] (example32) {$(x_6^{tr}, y_6^{tr})$};

\node[single node=lightpink, below=of example32] (test2) {$x_1^{te}$};

\node[bigbox, fit=(instruction2) (example12) (example22) (example32) (test2)] (box2) {};
\end{pgfonlayer}

\begin{pgfonlayer}{layer2}
\node[bigbox opaque, fit=(instruction2) (example12) (example22) (example32) (test2)] (box2) {};
\end{pgfonlayer}

\begin{pgfonlayer}{layer3}
\node[single node=lightyellow, below=-0.25cm of instruction2, xshift=0.25cm] (instruction3) {instruction};

\node[single node=lightblue, below=of instruction3] (example13) {$(x_1^{tr}, y_1^{tr})$};
\node[single node=lightblue, below=of example13] (example23) {$(x_2^{tr}, y_2^{tr})$};
\node[single node=lightblue, below=of example23] (example33) {$(x_3^{tr}, y_3^{tr})$};

\node[single node=lightpink, below=of example33] (test3) {$x_1^{test}$};

\node[bigbox, fit=(instruction3) (example13) (example23) (example33) (test3)] (box3) {};
\end{pgfonlayer}

\begin{pgfonlayer}{layer4}
\node[bigbox opaque, fit=(instruction3) (example13) (example23) (example33) (test3)] (box3) {};
\end{pgfonlayer}

\begin{pgfonlayer}{layer5}
\node[single node=lightyellow, below=-0.25cm of instruction3, xshift=0.25cm] (instruction4) {instruction};

\node[single node=lightblue, below=of instruction4] (example14) {$(x_1^{tr}, y_1^{tr})$};
\node[single node=lightblue, below=of example14] (example24) {$(x_2^{tr}, y_2^{tr})$};
\node[single node=lightblue, below=of example24] (example34) {$(x_3^{tr}, y_3^{tr})$};

\node[single node=lightpink, below=of example34] (test4) {$x_1^{test}$};

\node[bigbox, fit=(instruction4) (example14) (example24) (example34) (test4)] (box4) {};
\end{pgfonlayer}

\begin{pgfonlayer}{layer6}
\node[bigbox, fit=(instruction4) (example14) (example24) (example34) (test4)] (box4) {};
\end{pgfonlayer}

\draw [arrow] (box1.east) -- (box2.west);

\end{tikzpicture}

%% file: figures/permutation_example_tikz.tex
\begin{tikzpicture}[
  node distance = 0.2cm and 0.2cm,
  single node/.style = {rectangle, fill=#1, text opacity=1, opacity=0.5, align=center, font=\footnotesize, text width=2cm, inner sep=1.5mm, rounded corners=0.5mm},
  bigbox/.style = {draw, thick, rounded corners, rectangle},
  arrow/.style={thick, -latex},
  bigbox opaque/.style = {draw, thick, rounded corners, rectangle, opacity=1, fill=white},
  arrow/.style={thick, -latex},
]

\node[single node=lightyellow] (instruction) {instruction};

\node[single node=lightblue, below=of instruction] (example1) {$(x_1^{tr}, y_1^{tr})$};
\node[single node=lightblue, below=of example1] (example2) {$(x_2^{tr}, y_2^{tr})$};
\node[single node=lightblue, below=of example2] (example3) {$(x_3^{tr}, y_3^{tr})$};

\node[single node=lightpink, below=of example3] (test) {$x_1^{te}$};

\node[bigbox, fit=(instruction) (example1) (example2) (example3) (test)] (box1) {};

\begin{pgfonlayer}{layer1}
\node[single node=lightyellow, right=1cm of instruction] (instruction2) {instruction};

\node[single node=lightblue, below=of instruction2] (example12) {$(x_3^{tr}, y_3^{tr})$};
\node[single node=lightblue, below=of example12] (example22) {$(x_1^{tr}, y_1^{tr})$};
\node[single node=lightblue, below=of example22] (example32) {$(x_2^{tr}, y_2^{tr})$};

\node[single node=lightpink, below=of example32] (test2) {$x_1^{te}$};

\node[bigbox, fit=(instruction2) (example12) (example22) (example32) (test2)] (box2) {};
\end{pgfonlayer}

\begin{pgfonlayer}{layer2}
\node[bigbox opaque, fit=(instruction2) (example12) (example22) (example32) (test2)] (box2) {};
\end{pgfonlayer}

\begin{pgfonlayer}{layer3}
\node[single node=lightyellow, below=-0.25cm of instruction2, xshift=0.25cm] (instruction3) {instruction};

\node[single node=lightblue, below=of instruction3] (example13) {$(x_1^{tr}, y_1^{tr})$};
\node[single node=lightblue, below=of example13] (example23) {$(x_2^{tr}, y_2^{tr})$};
\node[single node=lightblue, below=of example23] (example33) {$(x_3^{tr}, y_3^{tr})$};

\node[single node=lightpink, below=of example33] (test3) {$x_1^{test}$};

\node[bigbox, fit=(instruction3) (example13) (example23) (example33) (test3)] (box3) {};
\end{pgfonlayer}

\begin{pgfonlayer}{layer4}
\node[bigbox opaque, fit=(instruction3) (example13) (example23) (example33) (test3)] (box3) {};
\end{pgfonlayer}

\begin{pgfonlayer}{layer5}
\node[single node=lightyellow, below=-0.25cm of instruction3, xshift=0.25cm] (instruction4) {instruction};

\node[single node=lightblue, below=of instruction4] (example14) {$(x_1^{tr}, y_1^{tr})$};
\node[single node=lightblue, below=of example14] (example24) {$(x_2^{tr}, y_2^{tr})$};
\node[single node=lightblue, below=of example24] (example34) {$(x_3^{tr}, y_3^{tr})$};

\node[single node=lightpink, below=of example34] (test4) {$x_1^{test}$};

\node[bigbox, fit=(instruction4) (example14) (example24) (example34) (test4)] (box4) {};
\end{pgfonlayer}

\begin{pgfonlayer}{layer6}
\node[bigbox, fit=(instruction4) (example14) (example24) (example34) (test4)] (box4) {};
\end{pgfonlayer}

\draw [arrow] (box1.east) -- (box2.west);

\end{tikzpicture}

%% file: tables/tasks.tex
\begin{table}[t]
\centering
\resizebox{\linewidth}{!}{
\begin{tabular}{ll}
\toprule
\textbf{Task Type} & \textbf{Tasks} \\ 
\midrule
\multirow{5}{*}{Classification (CLS)}      & AG News \cite{agnews} \\ 
                                            & ANLI \cite{anli} \\
                                            & BoolQ \cite{boolq} \\ 
                                            & IMDB \cite{imdb} \\ 
                                            & TweetEval Emotion \cite{emotion} \\
\midrule                                            
\multirow{2}{*}{Multiple-choice (MCQ)}                      & CosmosQA \cite{cosmosqa} \\ 
& HellaSwag \cite{hellaswag} \\
\midrule
\multirow{2}{*}{Generative QA (GQA)}                         & NQ-Open \cite{nqopen} \\ 
& TriviaQA \cite{triviaqa} \\ \bottomrule
\end{tabular}}
\caption{Tasks included in our evaluation suite.}
\label{table:tasks}
\end{table}

%% file: tables/models.tex
\begin{table}[t]
\centering
\resizebox{\linewidth}{!}{
\begin{tabular}{@{}ll@{}}
\toprule
\textbf{Model Family} & \textbf{Size} \\ \midrule
BLOOM~\cite{bloom}                 & 1.1B, 1.7B, 3B, 7.1B              \\
GPT Neo   ~\cite{gptneo, gptneox}            & 1.3B, 2.7B, 20B                   \\
LLaMA~\cite{llama}                 & 7B, 13B                           \\
OPT    ~\cite{opt}               & 1.3B, 2.7B, 6.7B, 13B             \\ \bottomrule
\end{tabular}}
\caption{Model families and corresponding model scales included in our evaluation suite.}
\label{table:models}
\end{table}

%% file: tables/methods.tex
\begin{table}[t] 
\centering
\resizebox{\linewidth}{!}{

\begin{tabular}{lcc}
\toprule 
\large \textbf{Method} & \large \textbf{Task-specific} & \large \textbf{Automatic induction} \\ 

\midrule
\large{Null instruction} & \xmark & \xmark\\
\large{Generic instruction} & \xmark & \xmark\\
\midrule
\large{PromptSource} \cite{promptsource} & \cmark & \xmark\\
\large{Ad hoc} & \cmark & \xmark \\
\midrule
\large{Low Perplexity} \cite{lowperplexityprompts} & \cmark & \cmark\\
\large{APE} \cite{ape} & \cmark & \cmark\\ 
\large{RLPrompt} \cite{rlprompt} & \cmark & \cmark\\
\bottomrule
\end{tabular}}
\caption{Instruction selection methods we evaluate}
\label{table:methods}
\end{table}

%% file: sections/experimental_setup.tex
\section{Experimental setup}
\label{sec:experimental_setup}
\label{sec:results}
We perform experiments evaluating 3 families of instruction selection methods (listed in \tableref{table:methods}). 

\paragraph{Task-agnostic instructions}
In practical ICL settings, it is straightforward to use instructions that contain no task-specific information.
\begin{itemize}[leftmargin=*]
    \item \textbf{Null instruction:} We assess the impact of omitting instructions from the prompt. This amounts to constructing prompts that consist of demonstrations and a test example in few-shot, and only an unanswered test-example in zero-shot settings.
    
    \item \textbf{Generic instructions:} We assess the impact of using generic task-agnostic instructions such as \texttt{Complete the following task:}. These instructions require minimal effort to write since they do not demand knowledge of the task. We list the set of generic instructions we evaluate in Table~\ref{table:generics}.
\end{itemize}

\paragraph{Manual task-specific instructions} 

We evaluate manually-written task-specific instructions that ICL practitioners may use in practice.

\begin{itemize}[leftmargin=*]
    \item \textbf{PromptSource:} PromptSource \cite{promptsource} is a public collection of manually-curated prompt templates pertaining to 170+ datasets which are often used off-the-shelf for ICL and are generally considered high-quality.
    
    \item \textbf{Ad hoc:} 
    ICL practitioners often create task-specific instructions ad hoc, based on the semantics of the given task. We simulate this mode of instruction selection by asking ChatGPT to generate several paraphrases of task-specific seed instructions we obtain from PromptSource and randomly sampling from the generated set.
    
\end{itemize}

\paragraph{Automatically synthesized task-specific instructions} We evaluate 3 popular automated instruction selection and induction methods that are representative of previous work.

\begin{itemize}[leftmargin=*]
    \item \textbf{Low Perplexity:} \cite{lowperplexityprompts} find that the perplexity a model associates with an instruction is negatively correlated with its ICL performance when using that instruction. We use the SPELL algorithm proposed by \citet{lowperplexityprompts} to select the least perplexity instructions (for each model) from a large pool of ChatGPT paraphrased instructions.

    \item \textbf{APE}: \cite{ape} is an automatic few-shot method for inducing instructions by prompting a language model to describe the given task, and refining the set of generated prompts using accuracy on a small held-out validation set. While \citet{ape} limit their evaluation to GPT-3~\cite{brown2020language} and InstructGPT~\cite{instructgpt}, we assess APE's applicability to a significantly larger set of models and tasks.

    \item \textbf{RLPrompt}~\cite{rlprompt} is a reinforcement-learning-based approach for few-shot prompt induction. While the original authors only evaluate their method using GPT-2 on a few classification tasks, we expand this assessment to many more models and tasks. Notably, we assess the extensibility of RLPrompt to MCQ tasks, but do not test RLPrompt performance on GQA tasks since the algorithm is not directly applicable to generation tasks. 
\end{itemize}

%% file: sections/results.tex
\section{Results}
\input{tables/accuracy_perc}

We tabulate the mean relative gain values over accuracy metrics in Table~\ref{table:accuracy_perc}, and the mean standard deviations corresponding to selectional and permutational sensitivity metrics in Table~\ref{table:std_mean}.

\subsection{Less sophisticated instruction selection methods tend to show higher accuracy}

We find that \textbf{task-agnostic instructions dominate in few-shot settings} with Null instructions and Generic instructions achieving the highest aggregated performance in 5/9 tasks for few-shot accuracy and 6/9 tasks for perturbation accuracy. Although both these methods show above-average performance in few-shot settings, Null instructions tend to perform better among the two.

Although PromptSource instructions only show an average performance in few-shot settings, their \textbf{manually curated task-specific instructions prove most effective in zero-shot settings}, achieving the highest aggregated performance in 6/9 tasks and usually achieving markedly higher mean relative gain values than even the runner-up method for the task. 
This is especially true of GQA tasks where PromptSource instructions outperform the average by $>$17\%.

\textbf{Automatic task-specific instructions are usually outperformed by simple baselines.} They fail to achieve the best zero-shot performance on any task we consider. While they do sometimes perform competitively with simpler baselines in the few-shot setting, emerging as the best-performing instructions in 2/9 tasks, this behavior is inconsistent. Although Low Perplexity instructions and APE instructions seldom show above-average performance in either setting, RLPrompt instructions show above-average performance in 5/7 tasks in both settings. They are still usually outperformed by instructions obtained through simpler means such as Null and PromptSource instructions.

\subsection{Ranges of variation of aggregated scores}
We notice that instructions have a more significant impact in zero-shot settings as compared to few-shot settings. For most tasks, we find that the highest mean relative gain values achieved in the zero-shot setting are markedly greater than those in the few-shot setting. Accordingly, the minimum values for each task are also relatively lower in zero-shot settings. This finding suggests that instructions play a significant role in informing models of semantics in zero-shot settings whereas in few-shot settings, most of a model's understanding of task-semantics comes from the demonstrations.

The degree of variation in accuracy due to instruction choice varies considerably across tasks. AG News and Emotion show the highest variability in few-shot performance while GQA tasks show the most variability in zero-shot settings.

\input{tables/std_mean}
Table~\ref{table:std_mean} shows that selectional and permutational sensitivities vary dramatically across tasks even though they are  roughly consistent across all methods for a given task implying that all the methods we evaluate are comparable in sensitivity, which is unsurprising since none of them explicitly optimize for it. We also find that most methods show comparable, but usually lower permutational sensitivity than selectional sensitivity across all tasks.

\subsection{Analysis}
\input{tables/accuracy_perc_scale_separated}
We tabulate the mean relative gain values for zero-shot and few-shot accuracies computed separately for ``small" models with $<6$ billion parameters and ``large"
 models with $\ge 6$ billion parameters in Table~\ref{table:accuracy_perc_scale_separated}. For ease of comparison, we average the mean relative gain values thus obtained by task-type. 
Although the observations that PromptSource and task-agnostic instructions tend to perform the best across zero- and few-shot settings persist across model scales, we find that the ranges of variation in the few-shot mean relative gain values for large models are consistently smaller than those for small models for every task-type. This suggests that large models are able to grasp task semantics from demonstrations (when provided) while small models are more sensitive to the instruction used. 

\subsection{Discussion}
Our findings reveal that in practical in-context learning settings, simpler prompting methods, such as task-agnostic or expert manually written instructions, often outperform automatically synthesized ones at the model scales we consider. Task-agnostic methods show strong performance in few-shot settings, whereas expert manual instructions appear crucial for achieving good zero-shot accuracy. The superiority of these straightforward methods over automatically induced instructions, which are often not competitive even with simple baselines, suggests a lack of transferability and generalizability among automatic induction methods. The competitive performance of automatic induction methods like APE and RLPrompt as reported by their authors implies either a limitation in their generalizability to a broader range of models and tasks, or the need for substantial hyperparameter tuning to get them to work well across models and tasks.

Our findings suggest that ICL practitioners may often be better off forgoing computationally expensive instruction induction or selection methods in favor of task-agnostic or manually written instructions, which seem to generalize better. Interestingly, we also find that methods that excel for one model and task do not necessarily also perform well for other tasks and models. Consequently, ICL practitioners may be forced to experiment with various instruction selection methods on a model- and task-specific basis in a manner reminiscent of hyperparameter tuning to find the best choice.

On the other hand, since few-shot ICL performance remains largely consistent regardless of the choice of instruction, practitioners could perhaps benefit from simply providing a few in-context demonstrations when available. The fact that null instructions tend to outperform all other methods in our study in few-shot settings suggests that it can be challenging to find instructions that reliably inform diverse models about task semantics. When models fail to grasp the semantics signaled by instructions, these may simply serve as a source of noise, hence impairing ICL performance.

Our findings underscore a broader issue regarding the inconsistent and often insufficient evaluation of instruction selection and induction techniques. We call for more comprehensive evaluations in this space and encourage the use of our evaluation suite to facilitate this process. 

%% file: tables/accuracy_perc.tex
\begin{table*}[htbp] 
\centering
\resizebox{\textwidth}{!}{
\begin{tabular}{lrrrrrrrrrc}
\toprule
 & \multicolumn{5}{c}{\textbf{CLS}} & \multicolumn{2}{c}{\textbf{MCQ}} & \multicolumn{2}{c}{\textbf{GQA}} & \multirow{2}{*}{\textbf{\# wins}}\\
\cmidrule(lr){2-6} \cmidrule(lr){7-8} \cmidrule(lr){9-10}

\textbf{Method} &  \textbf{AG News} &    \textbf{ANLI} &   \textbf{BoolQ} &    \textbf{IMDB} & \textbf{Emotion} & \textbf{HellaSwag} & \textbf{CosmosQA} & \textbf{TriviaQA} & \textbf{NQ-Open} & \\
\midrule
\rowcolor{gray!20} & \multicolumn{10}{c}{\bf{Zero-shot accuracy (mean relative gain) $\uparrow$}} \\ 
\midrule
Null Instruction    &   $2.26$ &  $1.07$ &  $\mathbf{2.48}$ & $-3.52$ & $-5.30$ &    $\mathbf{2.54}$ &   $\mathbf{5.94}$ &  $-3.08$ & $-25.67$ & 3\\
Generic Instruction &   $3.55$ & $-0.39$ &  $0.03$ &  $1.69$ &  $2.39$ &   $-0.13$ &  $-1.67$ &  $-1.52$ &  $-5.99$ & 0\\
\greyrule
PromptSource        &   $\mathbf{5.81}$ &  $\mathbf{1.38}$ & $-0.65$ &  $\mathbf{4.34}$ &  $\mathbf{5.13}$ &   $-1.54$ &  $-3.42$ &  $\mathbf{17.02}$ &  $\mathbf{22.15}$ & 6\\
Ad hoc              &  $-0.33$ &  $0.21$ &  $0.55$ &  $1.41$ &  $0.66$ &   $-0.27$ &  $-2.46$ &  $-2.03$ &   $2.31$ & 0\\
\greyrule
Low Perplexity      &  $-0.59$ &  $1.22$ &  $0.56$ &  $0.84$ & $-4.07$ &   $-1.38$ &  $-2.18$ &  $-5.87$ &   $2.81$ & 0\\
APE                 & $-15.63$ & $-3.86$ & $-1.07$ & $-1.77$ & $-0.26$ &   $-1.06$ &   $0.00$ &  $-4.70$ &   $4.39$ & 0\\
RLPrompt            &   $4.92$ &  $0.37$ & $-1.89$ & $-2.99$ &  $1.46$ &    $1.85$ &   $3.79$ &      $-$ &      $-$ & 0\\
\midrule
\rowcolor{gray!20} & \multicolumn{10}{c}{\bf{Few-shot accuracy (mean relative gain) $\uparrow$}} \\ 
\midrule
Null Instruction    &   $4.09$ & $-0.22$ &  $\mathbf{0.87}$ & $-0.80$ &  $\mathbf{5.89}$ &    $0.17$ &   $\mathbf{1.33}$ &   $\mathbf{0.45}$ & $-0.02$ & 4\\
Generic Instruction &   $\mathbf{5.16}$ & $-0.20$ & $-0.10$ &  $0.45$ &  $4.84$ &    $0.04$ &  $-0.18$ &   $0.11$ &  $0.11$ & 1\\
\greyrule
PromptSource        &   $0.83$ &  $0.14$ & $-0.79$ &  $0.39$ & $-4.39$ &   $-0.06$ &  $-0.94$ &  $-0.36$ &  $\mathbf{0.61}$ & 1\\
Ad hoc              &   $2.18$ & $-0.10$ & $-0.05$ &  $\mathbf{0.60}$ & $-5.63$ &   $-0.21$ &  $-0.59$ &   $0.09$ & $-0.49$ & 1\\
\greyrule
Low Perplexity      &  $-1.96$ &  $\mathbf{0.31}$ & $-0.40$ &  $0.20$ & $-6.79$ &   $-0.23$ &  $-0.61$ &  $-0.06$ & $-0.02$ & 1\\
APE                 & $-15.43$ &  $0.10$ &  $0.06$ & $-0.69$ &  $1.17$ &    $0.02$ &   $0.17$ &  $-0.24$ & $-0.19$ & 0\\
RLPrompt            &   $5.13$ & $-0.02$ &  $0.40$ & $-0.14$ &  $4.90$ &    $\mathbf{0.27}$ &   $0.81$ &      $-$ &     $-$ & 1\\
\midrule
\rowcolor{gray!20} & \multicolumn{10}{c}{\bf{Few-shot perturbation accuracy (mean relative gain) $\uparrow$}} \\ 
\midrule
Null Instruction    &   $4.09$ & $-0.08$ &  $0.11$ & $-0.27$ &  $\mathbf{5.98}$ &    $0.11$ &   $\mathbf{1.10}$ &   $\mathbf{0.81}$ &  $\mathbf{1.28}$ & 4\\
Generic Instruction &   $\mathbf{5.15}$ & $-0.18$ & $-0.16$ &  $\mathbf{0.56}$ &  $4.23$ &   $-0.02$ &  $-0.02$ &   $0.08$ &  $0.10$ & 2\\
\greyrule
PromptSource        &   $1.14$ &  $0.27$ & $-0.02$ &  $0.33$ & $-3.92$ &    $0.06$ &  $-0.53$ &  $-0.65$ &  $0.04$ & 0\\
Ad hoc              &   $1.68$ &  $0.51$ & $-0.34$ &  $0.37$ & $-5.87$ &   $-0.08$ &  $-0.63$ &  $-0.28$ & $-0.61$ & 0\\
\greyrule
Low Perplexity      &  $-2.39$ &  $\mathbf{0.68}$ & $-0.12$ & $-0.20$ & $-6.61$ &   $-0.09$ &  $-0.66$ &  $-0.03$ & $-0.78$ & 1\\
APE                 & $-14.32$ & $-1.20$ &  $\mathbf{0.28}$ & $-0.82$ &  $1.26$ &   $-0.13$ &   $0.21$ &   $0.06$ & $-0.03$ & 1\\
RLPrompt            &   $4.65$ & $-0.01$ &  $0.24$ &  $0.03$ &  $4.94$ &    $\mathbf{0.15}$ &   $0.53$ &      $-$ &     $-$ & 1\\
\bottomrule
\end{tabular}
}
\caption{Mean relative gain values associated with zero-shot accuracy, and few-shot accuracy with unperturbed and perturbed test inputs. Only values that correspond to the same task and metric should be compared. Positive values represent above-average performance, and negative values represent below-average performance. The `\# wins' column shows the number of tasks where a method achieved the highest aggregated performance.
}
\label{table:accuracy_perc}
\end{table*}

%% file: tables/std_mean.tex
\begin{table*}[] 
\centering
\resizebox{\textwidth}{!}{
\begin{tabular}{lrrrrrrrrrc}
\toprule
 & \multicolumn{5}{c}{\textbf{CLS}} & \multicolumn{2}{c}{\textbf{MCQ}} & \multicolumn{2}{c}{\textbf{GQA}} & \multirow{2}{*}{\textbf{\# wins}}\\
\cmidrule(lr){2-6} \cmidrule(lr){7-8} \cmidrule(lr){9-10}

\textbf{Method} &  \textbf{AG News} &    \textbf{ANLI} &   \textbf{BoolQ} &    \textbf{IMDB} & \textbf{Emotion} & \textbf{HellaSwag} & \textbf{CosmosQA} & \textbf{TriviaQA} & \textbf{NQ-Open} & \\

\midrule
\rowcolor{gray!20} & \multicolumn{10}{c}{\bf{Selectional sensitivity (mean standard deviation) $\downarrow$}} \\ 
\midrule
Null Instruction    &  $\mathbf{6.69}$ & $2.45$ & $4.73$ & $\mathbf{5.28}$ &  $6.97$ &    $2.46$ &   $\mathbf{8.10}$ &   $2.59$ &  $2.28$ & 3 \\
Generic Instruction &  $6.87$ & $2.50$ & $4.76$ & $5.40$ &  $6.97$ &    $2.48$ &   $8.16$ &   $2.61$ &  $2.26$ & 0\\
\greyrule
PromptSource        &  $6.73$ & $2.26$ & $4.85$ & $5.37$ &  $6.43$ &    $2.43$ &   $8.26$ &   $\mathbf{2.59}$ &  $2.28$ & 1\\
Ad hoc              &  $6.95$ & $2.41$ & $\mathbf{4.62}$ & $5.38$ &  $6.34$ &    $2.42$ &   $8.20$ &   $2.65$ &  $2.37$ & 1\\
\greyrule
Low Perplexity      &  $7.07$ & $\mathbf{2.17}$ & $4.69$ & $5.64$ &  $\mathbf{6.25}$ &    $2.42$ &   $8.27$ &   $2.59$ &  $2.30$ & 2\\
APE                 &  $7.44$ & $2.98$ & $4.63$ & $5.70$ &  $6.67$ &    $2.43$ &   $8.16$ &   $2.65$ &  $\mathbf{2.21}$ & 1\\
RLPrompt            &  $6.76$ & $2.30$ & $4.79$ & $5.50$ &  $6.96$ &    $\mathbf{2.36}$ &   $8.16$ &      $-$ &     $-$ & 1\\
\midrule
\rowcolor{gray!20} & \multicolumn{10}{c}{\bf{Permutational sensitivity (mean standard deviation) $\downarrow$}} \\
\midrule
Null Instruction    &  $6.02$ & $\mathbf{1.99}$ & $3.82$ & $\mathbf{4.14}$ &  $5.48$ &    $1.12$ &   $1.87$ &   $1.52$ &  $1.28$ & 2\\
Generic Instruction &  $\mathbf{6.01}$ & $2.19$ & $3.89$ & $4.56$ &  $5.49$ &    $1.15$ &   $1.68$ &   $\mathbf{1.33}$ &  $1.22$ & 2\\
\greyrule
PromptSource        &  $6.06$ & $2.15$ & $3.61$ & $4.69$ &  $4.30$ &    $\mathbf{1.07}$ &   $1.67$ &   $1.47$ &  $\mathbf{1.17}$ & 2\\
Ad hoc              &  $6.10$ & $2.37$ & $3.77$ & $4.61$ &  $4.37$ &    $1.11$ &   $1.66$ &   $1.41$ &  $1.23$ & 0\\
\greyrule
Low Perplexity      &  $6.13$ & $2.24$ & $\mathbf{3.50}$ & $4.61$ &  $\mathbf{4.29}$ &    $1.13$ &   $1.69$ &   $1.46$ &  $1.27$ & 2\\
APE                 &  $6.14$ & $2.36$ & $3.69$ & $4.84$ &  $5.08$ &    $1.10$ &   $1.78$ &   $1.41$ &  $1.21$ & 0\\
RLPrompt            &  $6.26$ & $2.06$ & $3.82$ & $4.89$ &  $5.64$ &    $1.08$ &   $\mathbf{1.65}$ &      $-$ &     $-$ & 1\\ 
\bottomrule
\end{tabular}
}
\caption{Mean standard deviation of few-shot accuracy on varying selections and permutations of demonstrations respectively. The `\# wins' column respresents the number of tasks where a method achieves best performance.
}
\label{table:std_mean}
\end{table*}

%% file: tables/accuracy_perc_scale_separated.tex
\begin{table}[t] 
\centering
\resizebox{\linewidth}{!}{
\begin{tabular}{lrrrrrr}
\toprule
{} & \multicolumn{3}{c}{ $\mathbf{<6}$\textbf{B parameters}} & \multicolumn{3}{c}{$\mathbf{\ge 6}$\textbf{B parameters}}  \\

\cmidrule(lr){2-4} \cmidrule(lr){5-7}
\textbf{Method} &     \textbf{CLS} &     \textbf{MCQ} &      \textbf{GQA} &                \textbf{CLS} &     \textbf{MCQ} &      \textbf{GQA} \\
\midrule
\rowcolor{gray!20} \multicolumn{7}{c}{\bf{Zero-shot accuracy (mean relative gain) $\uparrow$}} \\ 
\midrule
Null Instruction    & $-2.89$ &  $1.71$ & $-15.86$ &                            $2.07$ &  $\mathbf{7.19}$ & $-12.64$ \\
Generic Instruction & $1.71$ &  $0.69$ &  $-0.64$ &                             $1.16$ & $-2.76$ &  $-7.40$  \\
PromptSource        &  $\mathbf{2.77}$ & $-2.18$ &  $\mathbf{25.03}$ &           $\mathbf{3.70}$ & $-2.83$ &  $\mathbf{13.23}$\\
Ad hoc              &  $1.87$ & $-0.94$ &   $4.56$ &                            $-1.11$ & $-1.86$ &  $-5.01$ \\
Low Perplexity      & $-2.35$ & $-1.09$ &  $-8.24$ &                            $1.85$ & $-2.58$ &   $6.54$ \\
APE                 & $-3.13$ & $-0.54$ &  $-4.85$ &                            $-6.14$ & $-0.51$ &   $5.33$ \\
RLPrompt            &  $2.01$ &  $\mathbf{2.37}$ & $-$ &                        $-1.54$ &  $3.34$ &      $-$ \\
\midrule
Variation Range     & $5.90$ & $4.55$ & $40.89$ &                               $9.84$ & $10.02$ & $25.87$\\
\midrule
\rowcolor{gray!20} \multicolumn{7}{c}{\bf{Few-shot accuracy (mean relative gain) $\uparrow$}} \\ 
\midrule
Null Instruction    & $2.63$ &  $\mathbf{0.75}$ &  $\mathbf{0.89}$ &            $1.20$ &  $\mathbf{0.76}$ & $-0.57$ \\
Generic Instruction &  $\mathbf{3.09}$ & $-0.10$ & $-0.15$ &                    $0.80$ & $-0.03$ &  $0.41$ \\
PromptSource        &  $-1.18$ & $-0.58$ & $-0.20$ &                            $-0.28$ & $-0.41$ &  $\mathbf{0.51}$ \\
Ad hoc              &  $-0.55$ & $-0.45$ &  $0.04$ &                            $-0.65$ & $-0.35$ & $-0.47$ \\
Low Perplexity      & $-2.57$ & $-0.48$ & $-0.30$ &                             $-0.75$ & $-0.35$ &  $0.27$ \\
APE                 & $-4.10$ &  $0.13$ & $-0.28$ &                             $-1.62$ &  $0.06$ & $-0.13$ \\
RLPrompt            &  $2.69$ &  $0.73$ &     $-$ &                             $\mathbf{1.31}$ &  $0.32$ &     $-$ \\
\midrule
Variation Range     & $7.19$ & $1.33$ & $1.19$ & $2.93$ & $1.17$ & $1.08$ \\
\bottomrule
\end{tabular}}
\caption{Mean relative gain values for zero-shot and few-shot accuracy computed separately over models with $<6$ and $\ge 6$ billion parameters, and averaged by task-type. We also tabulate the total range of variation of these values in each setting.
}
\label{table:accuracy_perc_scale_separated}
\end{table}

%% file: sections/conclusion.tex
\section{Conclusion}
We conduct the broadest attempt to our knowledge, to systematically study the generalizability of popular instruction selection and induction methods for ICL in LLMs. We find that simpler approaches such as using task-agnostic instructions, expert manual instructions, or even omitting instructions entirely tend to show good performance more consistently when evaluating across a wide variety of tasks and models. Our work indicates the need for more systematic and consistent evaluations in the instruction induction space. To facilitate such analyses, we release the InstructEval suite which provides coverage over 13 diverse autoregressive LLMs and 9 tasks spanning classification, multiple-choice QA, and generative QA.

%% file: sections/appendix.tex
\appendix

\section{Variation across model families}
\input{tables/accuracy_family_separated}
\input{tables/templates}
\input{tables/instruction_examples}

We also tabulate the mean relative gain values for zero-shot and few-shot accuracies computed separately for each model family in Table~\ref{table:accuracy_family_separated}, to understand the effect that model family has on instruction performance. Although the trends we discuss in Section~\ref{sec:results} regarding task-agnostic instructions and PromptSource instructions tending to dominate few-shot and zero-shot settings persist, we note that the instruction selection method that emerges the best-performing alternative often changes on varying the choice of model family and task-type. For instance, the automatic instruction induction methods APE and RLPrompt do show above-average performance for certain model families and task-types, but this behavior does not consistently extend to other families and types as well. This indicates a lack of generalizability in these methods.

\section{Implementation details}
\subsection{Evaluation}


We ameliorate the effect of statistical noise by rerunning each instruction selection/induction method we study using 5 random seeds independently for every task (and for every model, where applicable) and report results for each instruction selection/induction method by averaging the aggregated scores associated with all 5 instructions. 

We use $K=6$ demonstrations randomly sampled from the task's training set for every experiment we perform in the few-shot setting.

To maintain consistency, we perform all our experiments using fixed task-specific prompt templates. Each prompt begins with the instruction being tested, and continues into a sequence annotated demonstrations and a test example, each of which follow the templates listed in Table~\ref{table:templates}.

\subsection{Instruction selection methods}
\label{sec:baseline-implementation-details}

\paragraph{PromptSource} We sample and evaluate a random subset of instructions from those included in the public PromptSource repository for each task in our evaluation suite.

\paragraph{Ad hoc} We obtain the set of ad hoc instructions we evaluate for a task by tasking ChatGPT with generating 40 paraphrases of instructions for the task that we obtain from PromptSource. We then select a random sample of instructions from this 40-instruction pool and perform evaluations using each sampled instruction.

\paragraph{Low Perplexity} For each task, we rerank a pool of ChatGPT paraphrases of PromptSource instructions using the SPELL algorithm described by \cite{lowperplexityprompts}. When prompting a specific model, we choose the instruction with the lowest perplexity as measured by that model. 

\paragraph{APE} We use the official repository released by \cite{ape} to generate instructions for each of the tasks we consider. To remain consistent with the original methodology, we use the OpenAI DaVinci to induce and evaluate instructions during the induction phase. We opt to use the simpler version of the methodology proposed by the authors since they report that the computationally intensive Monte-Carlo search strategy only provides marginal improvements in accuracy.

\paragraph{RLPrompt} We use the public repository released by \citet{rlprompt} to induce instructions for the RLPrompt baseline in our evaluations. Although the original work only performs evaluations over classification datasets with a fixed label-space, we augment the codebase to allow instruction induction for MCQ tasks as well by formulating these as cloze-style completion tasks. We create instructions for all tasks  using the default settings of hyperparameters included with the codebase.

\paragraph{Task-agnostic} We completely omit instructions from the prompt when evaluating null instructions. We list the set of generic instructions we evaluate in \tableref{table:generics}.
\input{tables/generics}

We include examples of the instructions we obtain for each method in Table~\ref{table:instruction_examples}.

\section{Drawbacks of aggregation techniques used in previous work}
\label{app:scoring}
Some previous works like the HELM \cite{helm} benchmark also face similar challenges when attempting to compare high-dimensional vectors -- each representing a model evaluated over a variety of tasks -- against each other. HELM resorts to scoring models using head-to-head win rates. The win rate associated with a model indicates the fraction of head-to-head comparisons between the given model and all other models, across all scenarios, where the given model performs better along a specific metric. A notable disadvantage of this scoring technique is that it obscures the magnitude of variation in the metric associated with each test model and only conveys ordinal information about the relative performances of each model. This characteristic of head-to-head win rates makes them unsuitable for spotting broad trends across families of prompting methods.

In other works like BIG-bench \cite{bigbench}, raw metric scores representing task performance are normalized to vary from a range of 0-100 such that a normalized score of 0 corresponds to poor performance, while a normalized score of 100 corresponds to excellent performance on the task. This is done in an attempt to be able to compare the performance of a model across a variety of tasks of varying difficulty such that the normalization proves more forgiving on difficult tasks. While this score does capture cardinal information associated with the underlying variable, it relies on the knowledge of human experts to determine raw score thresholds that constitute poor or excellent performance along a given metric. To apply such a normalization scheme in our case, one would need access to a large array of such threshold scores corresponding to each model scale, task, and metric we consider. Obtaining such threshold scores across all our settings is challenging given the number of tests we perform and the variety of metrics we consider. Hence, this type of normalization proves infeasible in our case.

%% file: tables/accuracy_family_separated.tex
\begin{table*}[t] 
\centering
\resizebox{\linewidth}{!}{
\begin{tabular}{lrrrcrrrcrrrcrrrc}
\toprule
{} & \multicolumn{4}{c}{\textbf{BLOOM}} & \multicolumn{4}{c}{\textbf{GPT Neo}} & \multicolumn{4}{c}{\textbf{LLaMA}} &  \multicolumn{4}{c}{\textbf{OPT}}\\

\cmidrule(lr){2-5} \cmidrule(lr){6-9} \cmidrule(lr){10-13} \cmidrule(lr){14-17}
\textbf{Method} &     \textbf{CLS} &     \textbf{MCQ} &      \textbf{GQA} & \textbf{\# wins} &                  \textbf{CLS} &     \textbf{MCQ} &      \textbf{GQA} & \textbf{\# wins} &                        \textbf{CLS} &     \textbf{MCQ} &      \textbf{GQA} & \textbf{\# wins} &                     \textbf{CLS} &     \textbf{MCQ} &      \textbf{GQA} & \textbf{\# wins} \\
\midrule
\rowcolor{gray!20} & \multicolumn{16}{c}{\bf{Zero-shot accuracy (mean relative gain) $\uparrow$}} \\ 
\midrule
Null Instruction    & $-1.40$ & $\mathbf{3.60}$ & $-11.93$ & 3 &           $-1.80$ & $\mathbf{1.34}$ & $-10.46$ & 1 &         $4.02$ & $\mathbf{9.25}$ & $-9.73$ & 3 &              $-1.22$ & $\mathbf{4.54}$ & $-22.07$ & 4 \\
Generic Instruction & $\mathbf{5.03}$ & $-1.27$ & $-0.72$ & 1 &            $-0.35$ & $-0.20$ & $-1.86$ & 1 &                    $-2.73$ & $-2.09$ & $-13.25$ & 0 &                   $1.33$ & $-0.47$ & $-3.47$ & 0\\
\greyrule
PromptSource        & $2.03$ & $-2.89$ & $\mathbf{14.22}$ & 2 &            $1.61$ & $-0.69$ & $\mathbf{35.75}$ & 2 &          $\mathbf{10.01}$ & $-3.89$ & $7.82$ & 2 &               $\mathbf{2.16}$ & $-2.70$ & $\mathbf{18.70}$ & 4\\
Ad hoc              & $0.45$ & $-1.15$ & $2.93$ & 0 &                      $\mathbf{2.23}$ & $0.56$ & $0.08$ & 2 &            $-3.70$ & $-2.94$ & $-2.56$ & 0 &                        $1.35$ & $-2.23$ & $-1.26$ & 0\\
\greyrule
Low Perplexity      & $-3.76$ & $-2.19$ & $3.94$ & 1 &                      $-2.85$ & $0.45$ & $-20.87$ & 1 &                      $4.97$ & $-4.87$ & $5.10$ & 2 &                         $2.09$ & $-1.50$ & $4.32$  & 1 \\
APE                 & $-6.10$ & $0.75$ & $-8.44$ & 0 &                     $0.14$ & $-2.00$ & $-2.87$ & 1                           & $-7.01$ & $-0.09$ & $\mathbf{12.62}$ & 2 &            $-5.18$ & $-0.92$ & $3.78$ & 0\\
RLPrompt            & $3.76$ & $3.15$ & $-$ & 2                             & $1.02$ & $0.54$ & $-$ & 1 &                          $-5.57$ & $4.64$ & $-$ & 0 &                             $-0.53$ & $3.28$ & $-$ & 0\\
\midrule
\rowcolor{gray!20} & \multicolumn{16}{c}{\bf{Few-shot accuracy (mean relative gain) $\uparrow$}} \\ 
\midrule
Null Instruction    &  $3.04$ &  $\mathbf{1.11}$ &  $\mathbf{0.85}$ &  4 &                     $\mathbf{1.31}$ &  $\mathbf{0.33}$ & $-0.11$ & 2 &                  $\mathbf{1.40}$ &  $0.79$ & $-0.35$ & 4 &                       $1.67$ &  $\mathbf{0.70}$ &  $0.11$ & 1\\
Generic Instruction &  $\mathbf{3.64}$ & $-0.24$ & $-0.44$ & 2 &                           $1.27$ &  $0.31$ &  $0.04$ & 2 &                                   $0.24$ & $-0.11$ &  $0.23$ & 2                                     &  $1.89$ & $-0.17$ &  $\mathbf{0.64}$ & 3\\
\greyrule
PromptSource        & $-1.21$ & $-0.83$ &  $0.19$ & 1                                          & $-0.44$ & $-0.17$ & $-0.08$ & 2                                    & $-0.30$ & $-0.65$ &  $0.18$ & 0                                       & $-0.79$ & $-0.34$ &  $0.20$ & 1\\
Ad hoc              & $-0.35$ & $-0.57$ &  $0.41$ & 1                                        & $-1.02$ & $-0.12$ & $-0.59$ & 0 &                                $-1.26$ & $-0.62$ &  $\mathbf{0.31}$ & 0                                  & $-0.20$ & $-0.33$ & $-0.77$ & 0\\
\greyrule
Low Perplexity      & $-3.00$ & $-0.58$ & $-0.23$ & 0                                           & $-1.04$ & $-0.15$ & $-0.15$ & 0                                  & $-0.94$ & $-0.59$ &  $0.25$ & 1                                      & $-1.37$ & $-0.38$ &  $0.07$ & 1\\
APE                 & $-5.26$ &  $0.29$ & $-0.78$ & 0                                          & $-1.14$ & $-0.29$ &  $\mathbf{0.85}$ & 2                          &  $0.29$ &  $0.38$ & $-0.62$ & 1                                              & $-3.64$ &  $0.05$ & $-0.24$ & 0\\
RLPrompt            &  $3.14$ &  $0.82$ &     $-$ & 1                                          &  $1.06$ &  $0.10$ &     $-$ & 1                                 & $0.57$ &  $\mathbf{0.80}$ &     $-$ & 1                             &  $\mathbf{2.45}$ &  $0.47$ &     $-$ & 3\\
\bottomrule
\end{tabular}}
\caption{Mean relative gain values for zero-shot accuracy and few-shot accuracy computed separately over individual model families and averaged by task-type. Positive values represent above-average performance, and negative values represent below-average performance. We also tabulate the number of tasks where a method achieved highest aggregated performance in the `\# wins' column under every model family.}
\label{table:accuracy_family_separated}
\end{table*}

%% file: tables/templates.tex
\begin{table*}[t]
\centering
\resizebox{\linewidth}{!}{
\begin{tabular}{ll}
\toprule
\textbf{Tasks}  & \textbf{Demonstration template}     \\ 
\midrule
AG News \cite{agnews} & \texttt{News: }(text)\texttt{\textbackslash nCategory: }(label)\\ 
ANLI \cite{anli} & \texttt{Premise: }(premise)\texttt{\textbackslash nHypothetisis: }(hypothesis)\texttt{\textbackslash nRelation: }(label) \\
BoolQ \cite{boolq} & \texttt{Passage: }(passage)\texttt{\textbackslash nQuestion: }(question) \texttt{\textbackslash nAnswer: }(label) \\ 
IMDB \cite{imdb} & \texttt{Review: }(text)\texttt{\textbackslash nSentiment: }(label) \\ 
TweetEval Emotion \cite{emotion} & \texttt{Tweet: }(text)\texttt{\textbackslash nEmotion: }(label)\\
\midrule                                            
CosmosQA \cite{cosmosqa} & \texttt{Passage: }(context)\texttt{\textbackslash nQuestion: }(question)\texttt{\textbackslash nAnswer: }(answer) \\ 
HellaSwag \cite{hellaswag}  & \texttt{Sentence: }(ctx)\texttt{\textbackslash nAnswer: }(answer) \\
\midrule
NQ-Open \cite{nqopen} & \texttt{Question: }(question)\texttt{\textbackslash nAnswer: }(answer)\\ 
TriviaQA \cite{triviaqa} & \texttt{Question: }(question)\texttt{\textbackslash nAnswer: }(answer)  \\ \bottomrule
\end{tabular}}
\caption{Tasks included in our evaluation suite, and the demonstrations templates we use for each task.}
\label{table:templates}
\end{table*}

%% file: tables/instruction_examples.tex
\begin{table*}[t] 
\centering
\resizebox{\linewidth}{!}{

\begin{tabular}{ll}
\toprule 
\large \textbf{Method} & \large \textbf{Example instruction}\\ 

\midrule
\large{Null instruction} & (empty string)\\
\large{Generic instruction} & \texttt{Solve the following task:}\\
\midrule
\large{PromptSource} \cite{promptsource} & \texttt{What label best describes this news article?} \\
\large{Ad hoc} & \texttt{Which newspaper section is most likely to feature this news article?} \\
\midrule
\large{Low Perplexity} \cite{lowperplexityprompts} & \texttt{Which part of a newspaper do you think this article belongs to? World News,} \\ & \texttt{Sports, Business or Science and Technology?}\\
\large{APE} \cite{ape} & \texttt{classify each input into one of the following categories: World, U.S.,} \\ & \texttt{Business, Sci/Tech, or Sports.}\\
\large{RLPrompt} \cite{rlprompt} & \texttt{Tools undergradCam firmwareCam} \\
\bottomrule
\end{tabular}}
\caption{Example instructions obtained using each method for the AG News task}
\label{table:instruction_examples}
\end{table*}

%% file: tables/generics.tex
\begin{table}[t]
\centering
\resizebox{\linewidth}{!}{
\begin{tabular}{c}
\toprule
\textbf{Generic Instructions} \\
\midrule
 \texttt{Solve the following task:} \\ 
 \texttt{Find the answer below:} \\  
 \texttt{Complete the problem.} \\
 \texttt{Find the best solution to the question below:} \\
 \texttt{Complete the question below:} \\
 \bottomrule
\end{tabular}}
\caption{Sample generic instructions}
\label{table:generics}
\end{table}